%% file: acl_latex.tex
\pdfoutput=1

\documentclass[11pt]{article}

\usepackage{acl}

\usepackage{times}
\usepackage{latexsym}

\usepackage[T1]{fontenc}

\usepackage[utf8]{inputenc}

\usepackage{microtype}

\usepackage{hyperref}
\usepackage{url}
\usepackage{multirow,booktabs,makecell}
\usepackage{booktabs}
\usepackage{graphicx}
\usepackage{adjustbox}
\usepackage{color}
\usepackage{comment}
\usepackage[labelformat=simple]{subcaption}

\usepackage[T1]{fontenc}
\usepackage{enumitem}
\usepackage{mathabx}
\usepackage{bm}
\usepackage{soul}
\usepackage{pifont}
\usepackage{colortbl}
\usepackage{graphbox} 

\newcommand{\methodname}{DocLap}

\def\va{{\bm{a}}}
\def\vh{{\bm{h}}}
\def\vi{{\bm{i}}}
\def\vw{{\bm{w}}}
\def\gI{{\mathcal{I}}}

\title{Automatic Layout Planning for Visually-Rich Documents with Instruction-Following Models}

\author{
  Wanrong Zhu\textsuperscript{\P},
  Jennifer Healey\textsuperscript{\S},
  Ruiyi Zhang\textsuperscript{\S},
  William Yang Wang\textsuperscript{\P},
  Tong Sun\textsuperscript{\S}
  \\
  \textsuperscript{\P}UC Santa Barbara,
  \textsuperscript{\S}Adobe Research
  \\
  \texttt{\{wanrongzhu,william\}@cs.ucsb.edu},
  \texttt{\{jehealey,ruizhang,tsun\}@adobe.com}
}

\begin{document}
\maketitle

\input{sections/0-abstract}
\input{sections/1-introduction}

\input{sections/3-method}
\input{sections/4-setup}

\input{sections/5-results}
\input{sections/7-conclusion}

\bibliography{anthology,custom}
\bibliographystyle{acl_natbib}



\end{document}

%% file: sections/0-abstract.tex
\begin{abstract}

Recent advancements in instruction-following models have made user interactions with models more user-friendly and efficient, broadening their applicability. 
In graphic design, non-professional users often struggle to create visually appealing layouts due to limited skills and resources. 
In this work, we introduce a novel multimodal instruction-following framework for layout planning, allowing users to easily arrange visual elements into tailored layouts by specifying canvas size and design purpose, such as for book covers, posters, brochures, or menus. 
We developed three layout reasoning tasks to train the model in understanding and executing layout instructions. 
Experiments on two benchmarks show that our method not only simplifies the design process for non-professionals but also surpasses the performance of few-shot GPT-4V models, with mIoU higher by 12\% on Crello~\cite{Yamaguchi2021CanvasVAELT}. 
This progress highlights the potential of multimodal instruction-following models to automate and simplify the design process, providing an approachable solution for a wide range of design tasks on visually-rich documents.
\end{abstract}

%% file: sections/1-introduction.tex
\section{Introduction}

The creation of visually-rich documents (e.g., posters, brochures, book covers, digital advertisements, etc) using available visual components, poses a significant challenge for both professionals and amateurs in the design field. 
Central to this challenge is the task of arranging these components in an efficient and aesthetically pleasing manner, a process known to be both tedious and time-consuming.
Existing toolkits such as Adobe Express\footnote{\url{https://www.adobe.com/express/}}, Canva\footnote{\url{https://www.canva.com/}}, and PicsArt\footnote{\url{https://picsart.com/}}, 
usually provide fixed templates to users.
These templates, while useful, often fail to fully accommodate the varied and evolving design needs of users, thereby potentially limiting creative expression. 
Existing research on automatic layout planning~\citep{Hsu2023PosterLayoutAN,Yamaguchi2021CanvasVAELT,Inoue2023TowardsFM} often requires detailed annotations and poses addition constraints on fixed canvas ratios, thereby diminishing user-friendliness and adaptability.

\begin{figure}[t]
    \centering
    \includegraphics[width=\linewidth]{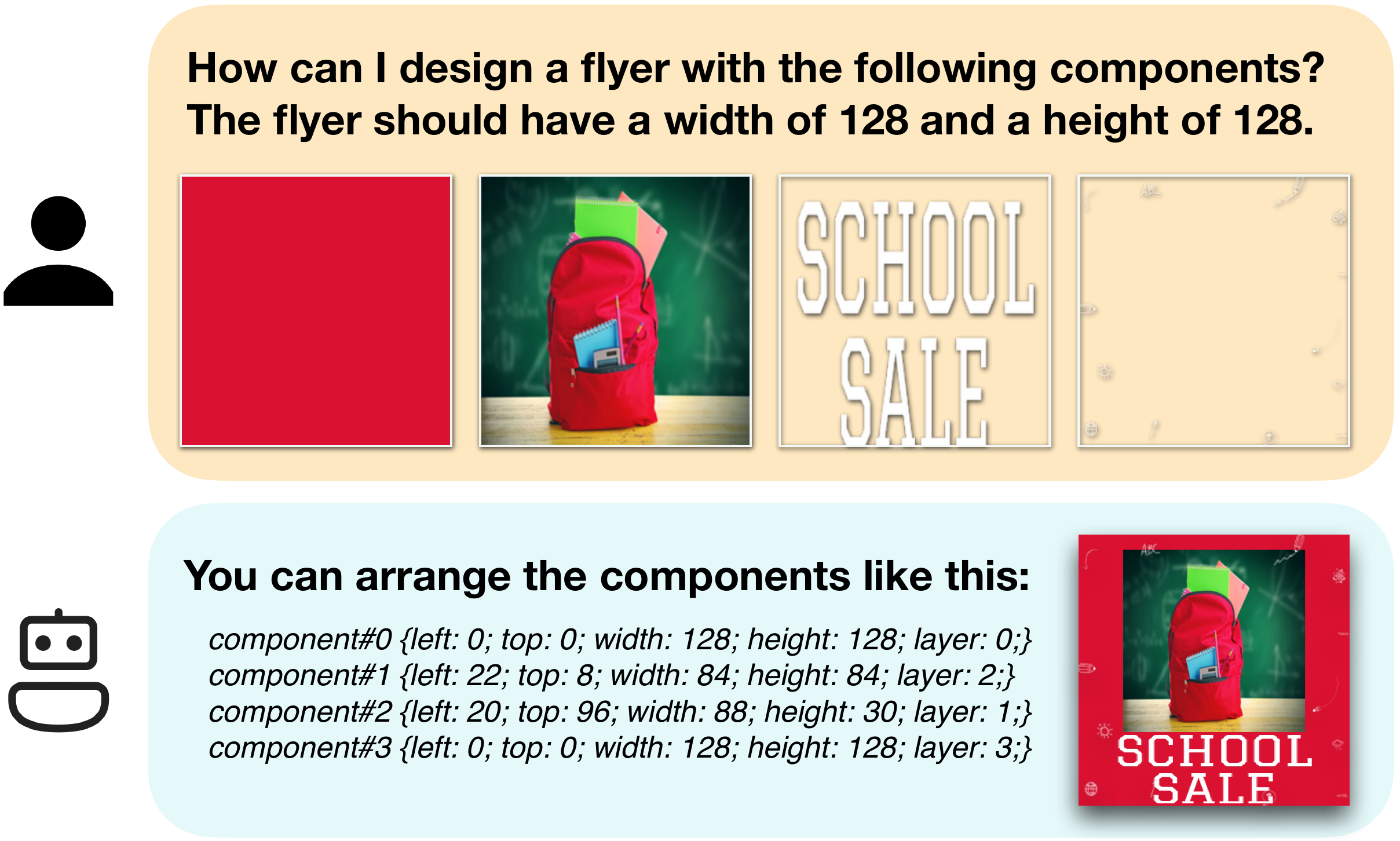}
\caption{
An example of a model conducting automatic layout planning following human-provided instructions and arranging visual contents for design purpose.
}
\label{fig:intro}
\end{figure}

\begin{figure*}[t]
    \centering
    \includegraphics[width=\linewidth]{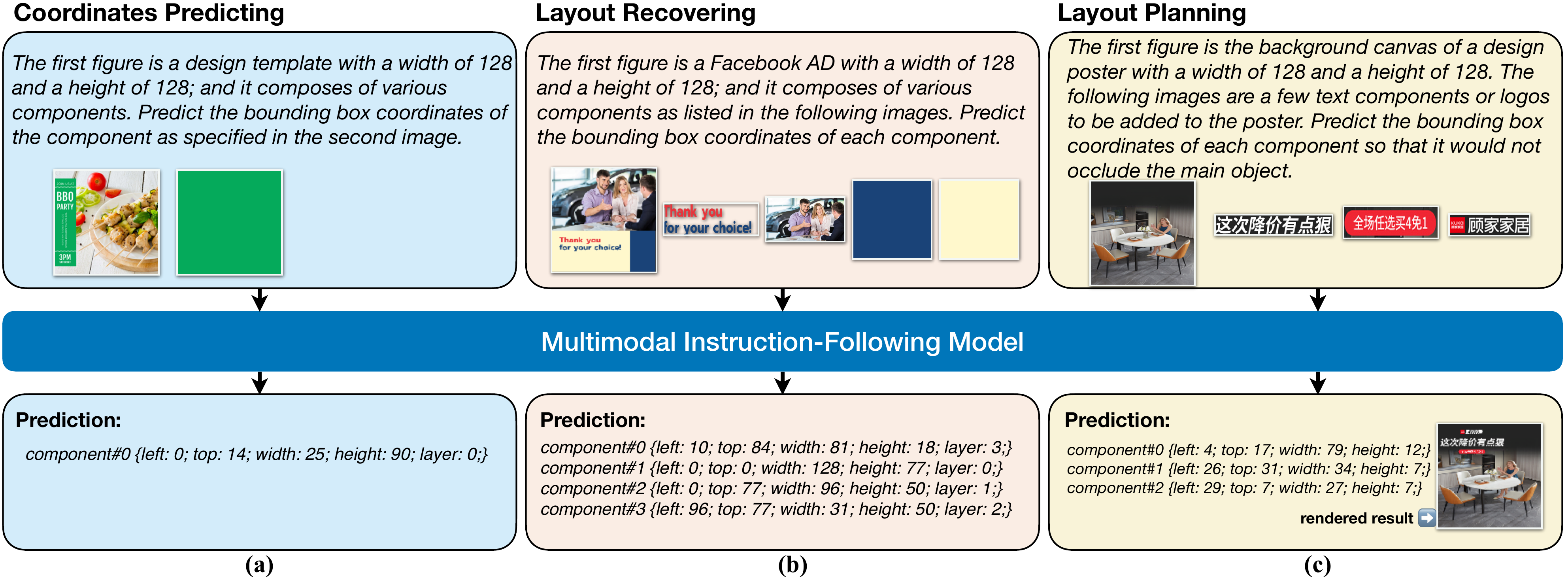}
\caption{Example inputs and outputs of the three layout reasoning tasks. (a) and (b) are examples from Crello~\citep{Yamaguchi2021CanvasVAELT}, while (c) is an example from PosterLayout~\citep{Hsu2023PosterLayoutAN}.
}
\label{fig:instruction_following_tasks}
\end{figure*}

Recent advancements in large language models (LLMs) have showcased their remarkable ability to follow human instructions and execute specified tasks~
\citep{Brown2020LanguageMA,Ouyang2022TrainingLM,openai2023gpt4}, introducing a new level of flexibility and control in human-computer interaction. Alongside these developments, we have witnessed the emergence of instruction-tuned multimodal models~\citep{Ye2023mPLUGOwlME,li2023mimicit,li2023otter,Awadalla2023OpenFlamingoAO,gpt4v}, extending the capabilities of LLMs to understand and process information across both textual and visual domains. This progression naturally raises the question of the potential application of instruction-following models in the complex domain of multimodal layout planning. However, employing these models for layout planning presents significant challenges, as the task requires intricate reasoning abilities, including but not limited to, cross-referencing multiple images and performing numerical calculations.

In this study, we propose \methodname, aiming to address the challenge of visually-rich \underline{doc}ument \underline{la}yout \underline{p}lanning using instruction-following models. 
To equip these models with the necessary knowledge beyond their primary focus on natural language processing, we have devised three instruction-following tasks focusing on layout reasoning. 
We evaluated our instruction-tuned \methodname~ model across two benchmark datasets, and the findings reveal that our approach not only succeeds in this novel application but also outperforms the baseline established by few-shot GPT-4(V). 
Our main contributions are:
\begin{itemize}[noitemsep, topsep=0.5pt,leftmargin=*]
\item We propose a novel method for solving the layout planning task using instruction-following models, opening new avenues for research in design automation.
\item We develop an instruction dataset featuring three layout reasoning tasks, aiming to enrich the resources available for future research.
\item Through experiments on two benchmark datasets, we validate the feasibility of our approach and demonstrate its competitive performance against few-shot GPT-4(V) models.
\end{itemize}

%% file: sections/3-method.tex
\section{Instruction-Guided Layout Planning for Visually-Rich Documents}

\paragraph{Task Definition}
Visually-rich documents consist of diverse design elements distributed across a canvas. To maintain the integrity of original text designs, text content is converted into images in our setup. The layout planning task involves arranging these design components, provided as a sequence of images ${\vi_1, \vi_2, ...\vi_n}$, where $n$ represents the component count, onto a canvas for specific application scenarios $\va$ (e.g., posters, Instagram posts, book covers) with defined dimensions $\vw$ (width) and $\vh$ (height). The canvas may either be blank or have a predefined background.

\paragraph{Instruction-Following Format}
To offer a more adaptable solution and enhance user experience, we approach this visually-rich layout planning task in an instruction-following manner~\citep{Ye2023mPLUGOwlME,li2023mimicit,li2023otter,Awadalla2023OpenFlamingoAO,gpt4v}. The model, in addition to receiving the sequence of design components ${\vi_1, \vi_2, ...\vi_n}$, will also be given instructions $\gI$ detailing the application scenarios $\va$ and the canvas size $(\vw, \vh)$. It is tasked with predicting the layout of each component in a structured format~\citep{feng2023layoutgpt,lin2023layoutprompter}. We adopt CSS to encapsulate layout properties including \texttt{top}, \texttt{left}, \texttt{width}, \texttt{height}, and another property \texttt{layer} that manages the stacking order of potentially overlapping elements.

\paragraph{Instruction-Following Format}
The task of layout planning encompasses challenges such as following instructions, cross-modal understanding, and numerical reasoning. To equip the model with essential knowledge, we designed three interrelated tasks, as illustrated in Figure~\ref{fig:instruction_following_tasks}: 
(a) \textit{Coordinates Predicting}, where the model predicts the coordinates of a specific component within a given design template; 
(b) \textit{Layout Recovering}, which involves predicting the coordinates of each component in a template given a sequence of components; 
and (c) \textit{Layout Planning}, where the model arranges a sequence of components on a canvas by predicting their coordinates. 
During preprocessing, components smaller than 5\% of the canvas size are excluded, and all templates are resized to ensure the longest edge does not exceed 128. While all three tasks contribute to model training, only the \textit{Layout Planning} task is evaluated during inference.

\input{tables/statistics}

\paragraph{Model}
\methodname~extends mPLUG-Owl~\citep{Ye2023mPLUGOwlME}, a multimodal framework integrating an LLM, a visual encoder, and a visual abstractor module. Specifically, it employs Llama-7b v1~\citep{Touvron2023LLaMAOA} as the LLM and CLIP ViT-L/14~\citep{Radford2021LearningTV} as the visual encoder. The visual abstractor module converts CLIP's visual features into 64 tokens that match the dimensionality of text embeddings, allowing for the simultaneous processing of multiple visual inputs. We extended the Llama v1 vocabulary with numerical tokens ranging from 0 to 128. The embeddings of the extended tokens are randomly initialized, and then tuned in further instruction tuning.

%% file: tables/statistics.tex
\begin{table}[t]
\begin{adjustbox}{width=\linewidth,center}
\begin{tabular}{llrrrr}\toprule
& &\textbf{Express} &\textbf{Crello} &\textbf{PosterLayout} \\\midrule
\multirow{3}{*}{Train} &Coordinates Predicting&581k &57k &26k \\
&Layout Recovering &160k &18k &9k \\
&Layout Planning &160k &18k &9k \\
\midrule
Val &Design Layout &- &1493 &591 \\
\bottomrule

\end{tabular}
\end{adjustbox}
\vspace{-7px}
\caption{
Number of examples contained in each training or validation tasks for the datasets used in this study.
}
\label{tab:data_statistics}
\end{table}

%% file: sections/4-setup.tex
\section{Experimental Setup}

\paragraph{Datasets}
We conduct experiments on layout planning for visually-rich documents with the following two benchmarks:
(1) Crello~\citep{Yamaguchi2021CanvasVAELT} is built upon design templates collected from online service. 
This task begins with an empty canvas, challenging the model to organize the layouts of the provided visual components.
(2) PosterLayout~\citep{Hsu2023PosterLayoutAN} starts from non-empty canvas (background image for posters), and requires the model to strategically place text, labels, and logos.
Our training data is supplemented with design templates from Adobe Express. 
Detailed dataset statistics are available in Table~\ref{tab:data_statistics}. To ensure fair comparison, validation examples are limited to no more than 4 images, aligning with the input constraints of GPT-4V at the time of our submission. Illustrative examples from both datasets are presented in Figure~\ref{fig:instruction_following_tasks}.

\paragraph{Baselines}
For Crello, we compare with CanvasVAE~\citep{Yamaguchi2021CanvasVAELT} and FlexDM~\citep{Inoue2023TowardsFM}.
For PosterLayout, we compare with  DS-GAN~\citep{Hsu2023PosterLayoutAN}.
Additionally, we include comparative evaluations with text-only versions of GPT-4 and GPT-4V~\citep{openai2023gpt4,gpt4v,gpt4vcontribution,gpt4vblog} across both tasks. For the text-only GPT-4 evaluations, visual components are not directly supplied. Instead, we employ BLIP-2~\citep{Li2023BLIP2BL}  to generate textual descriptions of each component.

\input{tables/crello_results}

\input{tables/posterlayout_results}

\paragraph{Metrics}
For Crello evaluation, we measure mean Intersection-over-Union (mIoU) between predicted and actual bounding boxes, along with accuracy in width, height, left, and top dimensions following FlexDM~\citep{Inoue2023TowardsFM}. Accuracy is quantified by assigning a score of 1 if the predicted value falls into the same 64-bin quantized range as the ground truth; otherwise, it scores 0.
In assessing PosterLayout, we follow DS-GAN~\citep{Hsu2023PosterLayoutAN} and employ content-aware metrics, including (1) occlusion rate$\downarrow$, indicating the percentage of primary objects obscured by design elements; (2) utility rate$\uparrow$, reflecting the extent to which design components cover non-primary object areas; and (3) unreadability$\downarrow$, measuring the uniformity of areas where text-containing elements are placed.

%% file: tables/crello_results.tex
\begin{table}[t]
\begin{adjustbox}{width=\linewidth,center}
\begin{tabular}{llrrrrrr}\toprule
&\textbf{Model} &\textbf{mIoU} &\textbf{Left} &\textbf{Top} &\textbf{Width} &\textbf{Height} \\\midrule
\texttt{\#1} &CanvasVAE & 42.39 &29.31 &30.97 &27.58 &29.99 \\
\texttt{\#2} &FlexDM &50.08 &34.98 &34.03 &30.04 &33.08 \\
\midrule
\texttt{\#3} &GPT-4 0-shot &30.75 &24.36 &24.07 &13.63 &15.11 \\
\texttt{\#4} &GPT-4 1-shot &29.97 &26.09 &23.71 &13.94 &13.33 \\
\texttt{\#5} &GPT-4V 0-shot &28.81 &19.96 &18.09 &10.45 &10.08 \\
\texttt{\#6} &GPT-4V 1-shot &35.17 &22.77 &20.90 &13.16 &14.11 \\
\midrule
\texttt{\#7} &\methodname~(Ours) &43.75 &33.46 &35.61 &19.18 &22.79 \\
\bottomrule
\end{tabular}
\end{adjustbox}
\vspace{-7px}
\caption{Automatic evaluation results on Crello showing mIoU and the accuracy for left, top, width and height.
}
\label{tab:crello_results}
\end{table}

%% file: tables/posterlayout_results.tex
\begin{table}[t]
\begin{adjustbox}{width=0.8\linewidth,center}
\scriptsize
\begin{tabular}{llrrrr}\toprule

&\textbf{Model} &\textbf{Occ.}$\downarrow$ &\textbf{Uti.}$\uparrow$ &\textbf{Rea.}$\downarrow$ \\
\midrule
\texttt{\#1} &DS-GAN &21.57 &23.92 &20.16 \\
\midrule
\texttt{\#2} &GPT-4 0-shot &50.61 &43.09 &25.87 \\
\texttt{\#3} &GPT-4 1-shot &47.92 &38.00 &25.34 \\
\texttt{\#4} &GPT-4V 0-shot &36.67 &33.26 &24.39 \\
\texttt{\#5} &GPT-4V 1-shot &36.39 &20.24 &26.03 \\
\midrule
\texttt{\#6} &\methodname~(Ours) &23.01 &22.46 &21.00 \\

\bottomrule
\end{tabular}
\end{adjustbox}
\vspace{-7px}
\caption{Evaluation results on PosterLayout.
\textit{Occ.}: occlusion rate; \textit{Uti.}: utility rate; \textit{Rea.}: unreadability.
}
\label{tab:posterlayout_results}
\end{table}

%% file: sections/5-results.tex
\section{Results \& Analysis}
\paragraph{Quantitative Results}

Table~\ref{tab:crello_results} shows the automatic evaluation results on Crello dataset. 
The first two lines are results from models that are trained with supervised learning. Line \texttt{\#3-\#6} show few-shot GPT-4(V) results, in which we notice that GPT-4V surpasses text-only GPT-4, and that providing demonstrative examples leads to better results compared to zero-shot prompting. Our \methodname's performance (\texttt{\#7}) surpass the few-shot GPT-4(V)  on both mIoU and aspect accuracies, but still falls behind a bit compared to FlexDM (\texttt{\#2}).

\begin{figure}[t]
    \centering
    \includegraphics[width=\linewidth]{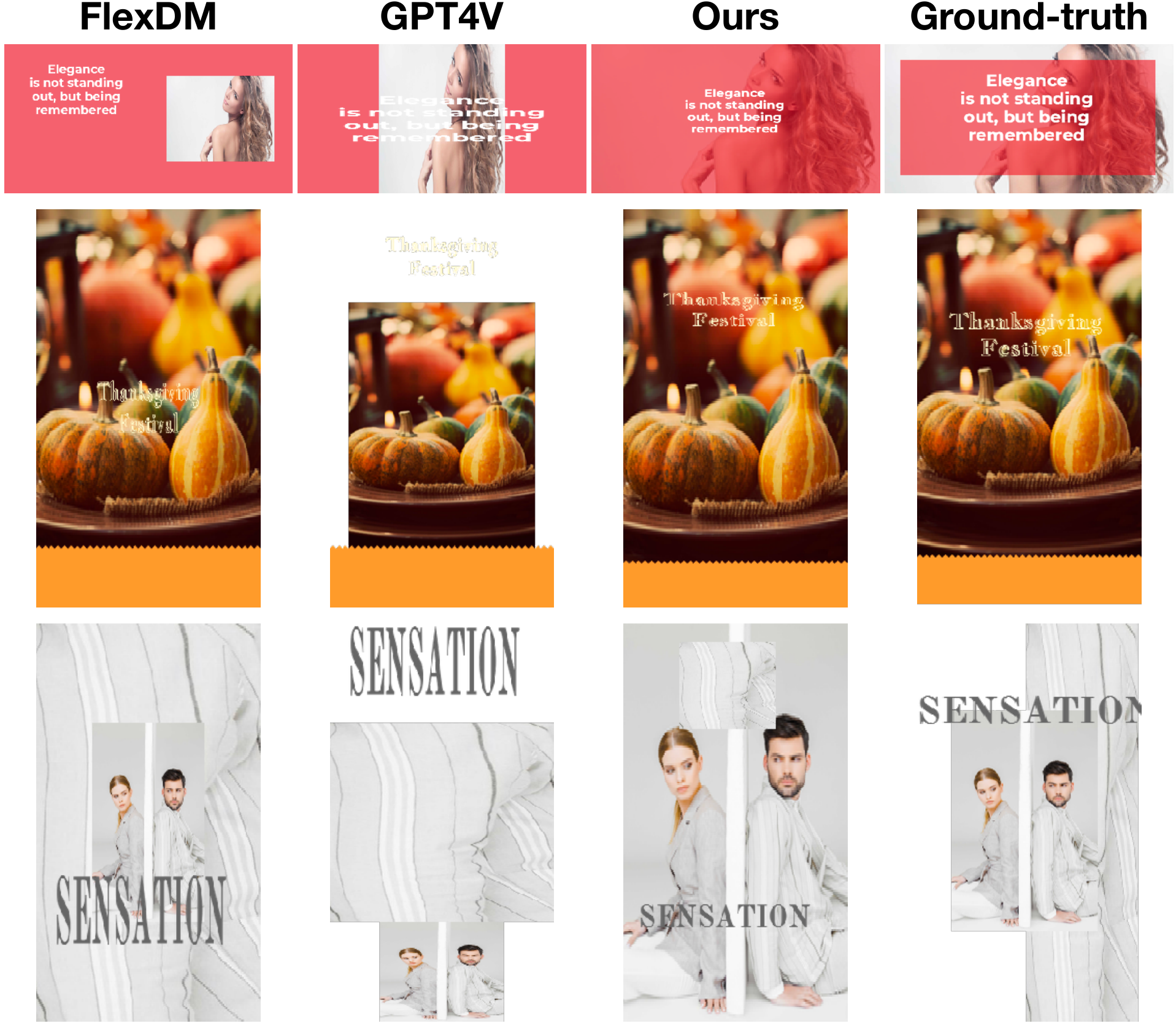}
\caption{
Qualitative comparisons for layout planning results on Crello.
GPT-4V w/ 1-shot learning.
}
\label{fig:crello_egs}
\end{figure}

\begin{figure}[t]
    \centering
    \includegraphics[width=0.95\linewidth]{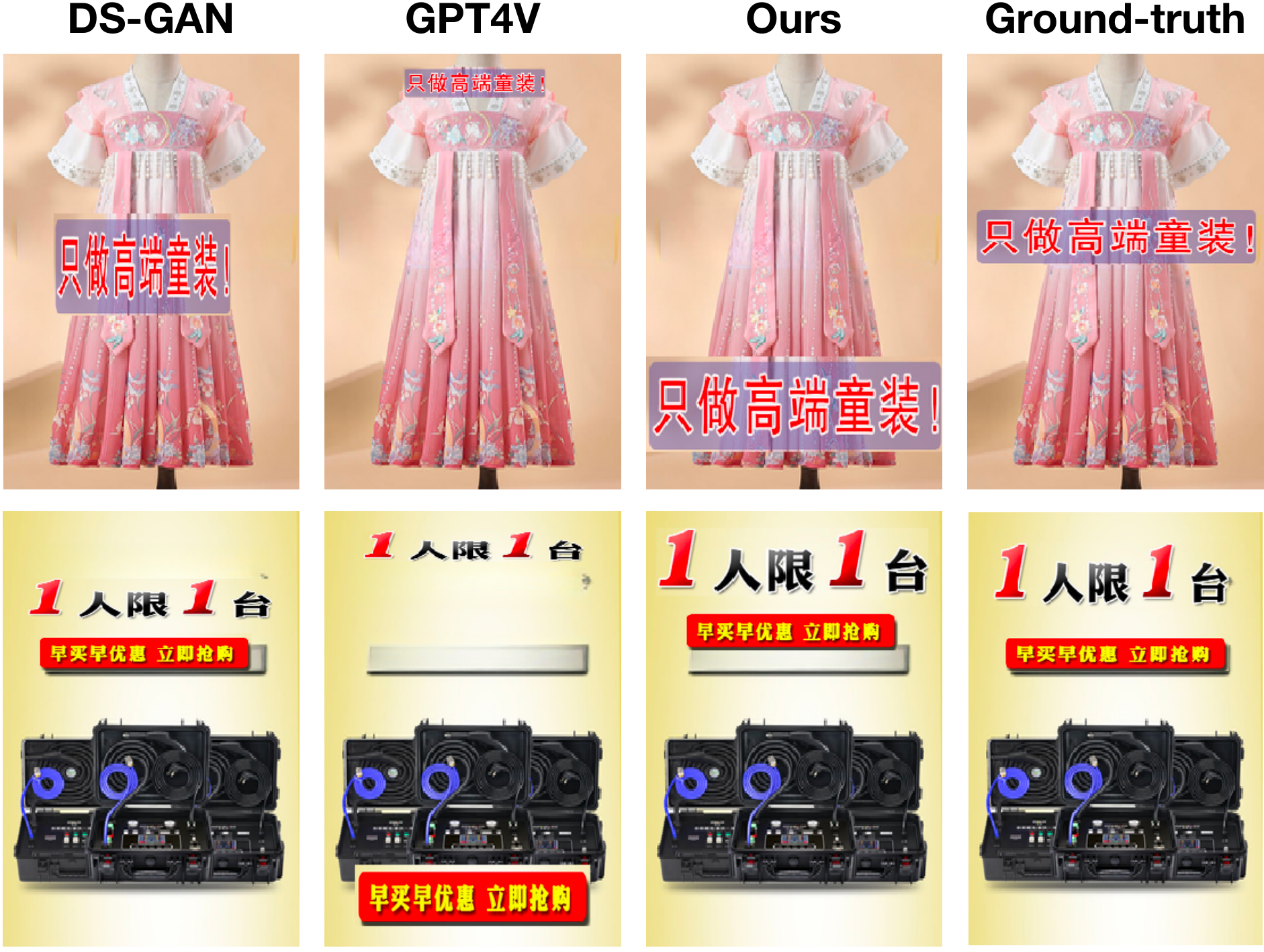}
\caption{
Qualitative comparisons for layout planning results on PosterLayout.
GPT-4V w/ 1-shot learning.
}
\label{fig:posterlayout_egs}
\end{figure}

Table~\ref{tab:posterlayout_results} presents the PosterLayout evaluation results, which reveals a trade-off between occlusion rate and utility rate across models. 
GPT-4(V) models (\texttt{\#2-\#5}) exhibit high occlusion and utility rates, indicating a propensity for predicting larger bounding boxes. Our \methodname~ shows a reduced occlusion rate, accompanied by a decrease in utility rate. 
Regarding unreadability, \methodname~ outperforms GPT-4(V), though DS-GAN (\texttt{\#1}) achieves the highest performance, underscoring the efficacy of supervised models in this context.
\begin{figure}[t]
    \centering
    \includegraphics[width=\linewidth]{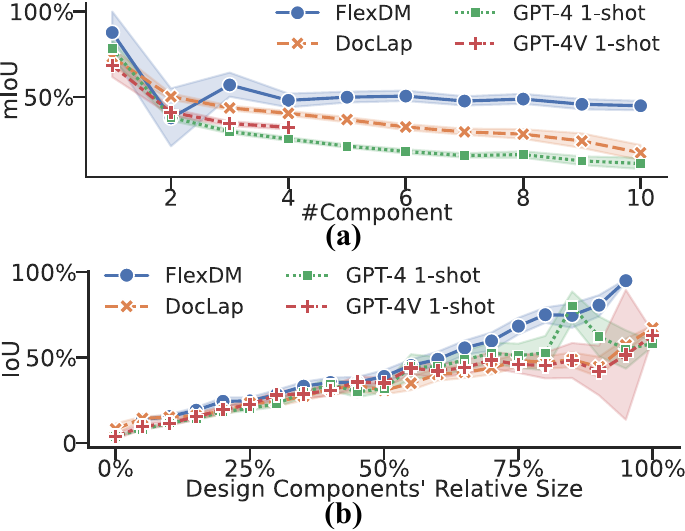}
\caption{
(a) mIoU variation with the number of visual components in design templates.
(b) IoU correlation with the relative size of a single visual component. Both plots pertain to Crello.
}
\vspace{-5px}
\label{fig:crello_analysis}
\end{figure}
\paragraph{Effects of \#Component}
Figure~\ref{fig:crello_analysis}(a) reveals that all listed models exhibit high mIoU for templates with a single component. FlexDM's mIoU shows slight fluctuations, stabilizing around 50\%. 
In contrast, mIoU for \methodname~ and GPT-4(V) decreases as the number of components increases, indicating that more complex scenarios involving more visual components might pose challenges to current instruction-following models.
%
\paragraph{Effects of Component Size}
Figure~\ref{fig:crello_analysis}(b) demonstrates a linear correlation between the relative size of a single visual component and the IoU of the model prediction with the ground truth for all models assessed. This suggests that smaller visual components pose a greater challenge for precise placement in accordance with the ground truth during layout planning. Typically, these small components, such as logos, small text boxes, or decorative elements, have a degree of positional flexibility, allowing for multiple valid placements.
\paragraph{Demonstrative Examples}
Figure~\ref{fig:crello_egs} shows examples from Crello while Figure~\ref{fig:posterlayout_egs} shows examples from PosterLayout.

%% file: sections/7-conclusion.tex
\section{Conclusion}
This study demonstrates the potential of instruction-following models in addressing the intricate task of layout planning for visually rich documents. 
The positive outcomes observed from our experiments on two distinct benchmarks affirm the viability and effectiveness of our methodology. 
This research paves the way for future explorations into the application of instruction-following models across various domains, highlighting their potential to revolutionize tasks that require a nuanced understanding of both language and visual elements.

\newpage

\section*{Limitations}
This study, while pioneering in its approach to simplifying the graphic design process through instruction-following models, acknowledges several limitations. First, the performance of our model, \methodname, and GPT-4(V) diminishes as the complexity of the layout increases, particularly with the addition of more visual components. This suggests a need for improved model robustness and adaptability in handling more intricate design scenarios.
Additionally, the evaluation metrics, such as mIoU and the binary accuracy measurement for layout attributes, may not fully capture the nuances of aesthetic and functional design quality. The reliance on these metrics might overlook the subjective and context-specific nature of effective design, indicating a potential area for developing more comprehensive evaluation frameworks.

\section*{Ethics Statement}
Our work on instruction-following models for layout planning, while innovative, introduces potential risks including over-reliance on automation, which may impede the development of design skills and creativity. Importantly, our model does not generate new visual content; all predictions are based on existing components provided by users. The outputs are solely layouts in text formats, mitigating risks related to copyright infringement and originality. However, the reliance on automated tools could lead to a homogenization of design aesthetics and potentially amplify biases present in the input data. Addressing these challenges requires careful consideration of the ethical implications of automated design tools and the promotion of responsible usage to complement human creativity.
Noted here that we utilize ChatGPT to polish the writing and ensure clarity and conciseness in the presentation of our research, without altering the fundamental nature of the work or its implications.